\documentclass{ifacconf}

\usepackage{graphicx}      
\usepackage{natbib}        
\usepackage{amsmath}
\begin{document}
\begin{frontmatter}

\title{Precision Landing of a UAV on a Moving Platform for Outdoor Applications} 

\author[First]{Adarsh Salagame} 
\author[First]{Sushant Govindraj} 
\author[First]{S. N. Omkar}

\address[First]{Department of Aerospace, 
   Indian Institute of Science, Bangalore (e-mail: adarshsalagame@gmail.com).}

\begin{abstract}                
As UAV technology improves, more uses have been found for these versatile autonomous vehicles, from surveillance to aerial photography, to package delivery, and each of these applications poses unique challenges. This paper implements a solution for one such challenge: To land on a moving target. This problem has been addressed before with varying degrees of success, however, most implementations focus on indoor applications. Outdoor poses greater challenges in the form of variables such as wind and lighting, and outdoor drones are heavier and more susceptible to inertial effects. Our approach is purely vision based, using a monocular camera and fiducial markers to localize the drone and a PID control to follow and land on the platform.
\end{abstract}


\end{frontmatter}

\section{Introduction}
As the technology for autonomy and aerial vehicles improves, so too does the demand for UAVs. Over the last few years, the field of Unmanned Aerial Vehicles has seen an upsurge of interest as more people find applications for the versatile and dynamic capabilities of drones. From aerial photography to package delivery to search and rescue in disaster management, drones have permeated many different sectors of industry and consumerism. However, the technology is not yet perfect and there are many problems to be solved before they become commonplace. 

One of these problems is landing on a moving target. Applications such as package delivery, dynamic docking stations, and envoy surveillance would benefit from the technology, and there have been multiple approaches to solving this problem. However, most of them rely on methods that would perform poorly in outdoor conditions. There are four primary challenges facing any potential solution.

\begin{enumerate}
	\item The target must be correctly identified and continuously tracked by the UAV for the duration of the landing procedure. 
	\item For vision-based landing, as the height is reduced, the field of view is shortened, increasing the risk of losing the target. 
	\item With outdoor applications, wind plays a role, forcing any system to be more robust to variations. 
	\item Outdoor applications often require payload capacity in the form of cameras, additional processors or packages to be carried, and the drones that carry them can be heavier than indoor drones. In such cases, inertia can affect the system and must be accounted for.
\end{enumerate}

\section{Previous Works}

There have been attempts at solving this problem, using various techniques. \cite{Wenzel2011} uses an infrared camera to track a pattern of infrared lights arranged on the platform, however, this method is only suitable for indoor use and would perform poorly in direct sunlight. \cite{Herisse2012} uses optical flow and a PI control to track a textured platform. \cite{Lee2012} uses image-based visual servoing to land on a platform with a single marker. This can produce noisy results when the system is highly nonlinear(\cite{10.1007/BFb0109663}), and reliance on a single marker means the system is more susceptible to variance and is more likely to lose the target. \cite{Kim2014} and \cite{Baca2017} both use fish eye lenses to solve for the field of view issue. Still, the first work uses color-based tracking, which does not provide enough specificity for outdoor use and the second work relies on prior knowledge about the expected path of the target.

The paper closest to the presented work is \cite{Araar2017}, which also uses fiducial markers and vision-based navigation for landing. However, the algorithms presented are implemented on the lightweight Parrot AR Drone under indoor conditions. The following work has been implemented and tested outdoors under challenging pre-monsoon windy conditions using the heavier Iris drone.

The next section details our approach and the steps we took to solve the four challenges presented above.

\section{Methodology}

\begin{figure}[h]
  \includegraphics[width=\linewidth]{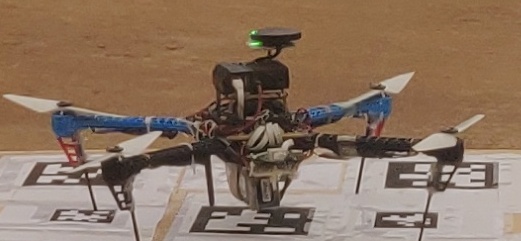}
  \caption{Setup of the drone}
  \label{fig:setup}
\end{figure}

Our setup uses a 3DR Iris drone with a Pixhawk 2.0 Cube Black flight controller unit. A downward-facing Logitech USB camera is mounted at the front of the drone for monocular vision, connected to an Odroid XU4 onboard to carry out all the processing, and it also carries a router for wireless connection to a ground station for monitoring. The total weight including the battery is 2.5kg. The setup is shown in Figure \ref{fig:setup}.

\subsection{The Landing Pad}

\begin{figure}[h]
\begin{center}
  \includegraphics[width=0.5\linewidth]{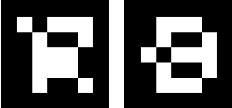}
  \caption{Examples of ArUco markers}
  \label{fig:aruco}
\end{center}
\end{figure}

In order to clearly differentiate the target from any other objects in the environment, fiducial markers are used. Fiducial markers are objects or patterns that can be used by an imaging system as a point of reference or measurement. They are designed with a preset size and unique ID which is identified by the imaging system to obtain data such as relative position and orientation to the marker in 3D space, enabling the drone to localize itself with respect to the target. Figure \ref{fig:aruco} shows one example of such markers, called ArUco markers. Other popular examples are Apriltags and ARTags.

\begin{figure}[h!]
  \includegraphics[width=\linewidth]{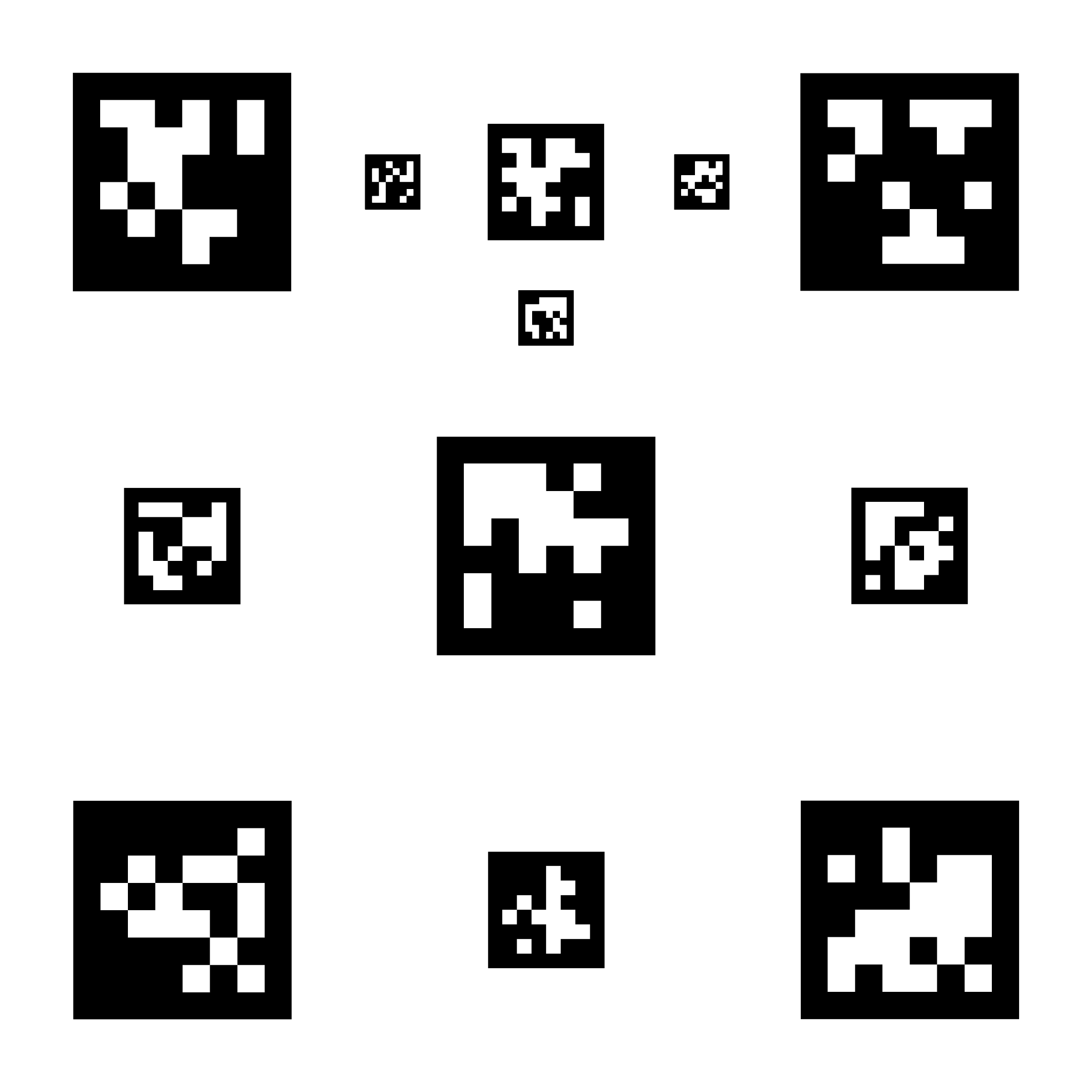}
  \caption{Design of markers on our landing platform}
  \label{fig:marker_board}
\end{figure}

Figure \ref{fig:marker_board} shows the pattern of markers used on our landing pad.  These are Apriltag Bundles. A bundle is a group of markers in a specific arrangement that collectively provide the same information as one smaller marker, making the measurement more sensitive and allowing for a wider range of detection at lower heights. In this work, three bundles of descending sizes were used to make sure at least one is detected at any point in the trajectory as the drone moves in for a landing. The largest bundle of five markers can be detected from a maximum height of 5m to a minimum height of 2m. When the camera is too close to capture the full marker within the frame, the next bundle of four markers and the smallest bundle are tracked until the drone lands.

\subsection{PID Control}

In order to follow the target, a PID controller is used. The Apriltag bundles allow the drone to localize itself with respect to the target, providing the error in position, ${\Delta x}$ and ${\Delta y}$, along the x and y axis of the drone. The controller is designed assuming the position error and the velocity effort to bridge that error is linearly correlated, such that when the drone is right above the marker, the velocity error is zero and the drone is effectively moving at the speed of the target.

\begin{align*}
	& {\Delta x} \texttt{ } {\displaystyle \propto} \texttt{ } {\Delta V_x}\\
	& {\Delta y} \texttt{ } {\displaystyle \propto} \texttt{ } {\Delta V_y}
\end{align*}

\begin{figure}[h]
  \includegraphics[width=\linewidth]{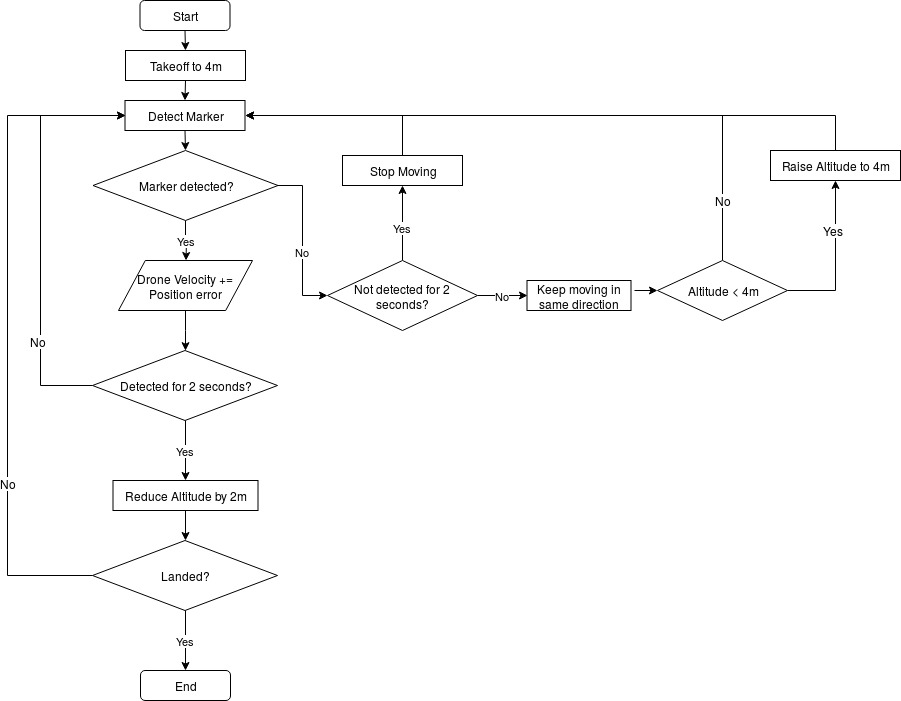}
  \caption{Flowchart of the autonomous landing algorithm}
  \label{fig:flowchart}
\end{figure}

The landing happens in two stages, going from a maximum height of 4m to 2m after continuously detecting the marker for 2 seconds, then initiates landing after continuously detecting it for two more seconds. If at any point, the marker detection is lost, the drone raises the altitude to increase the field of view. The full flowchart of the algorithm is shown in Figure \ref{fig:flowchart}.

\subsection{Interfacing with the drone}

Once the velocity effort along the x and y (pitch and roll) axis of the drone is obtained, these are fed to the drone as velocity setpoints. For this, mavros is used to interface between our onboard processor and the Pixhawk FCU present on the drone. Setpoints are sent using a Body NED Frame.


\section{Testing Methodology}

\begin{figure}[h!]
\begin{center}
  \includegraphics[width=0.7\linewidth]{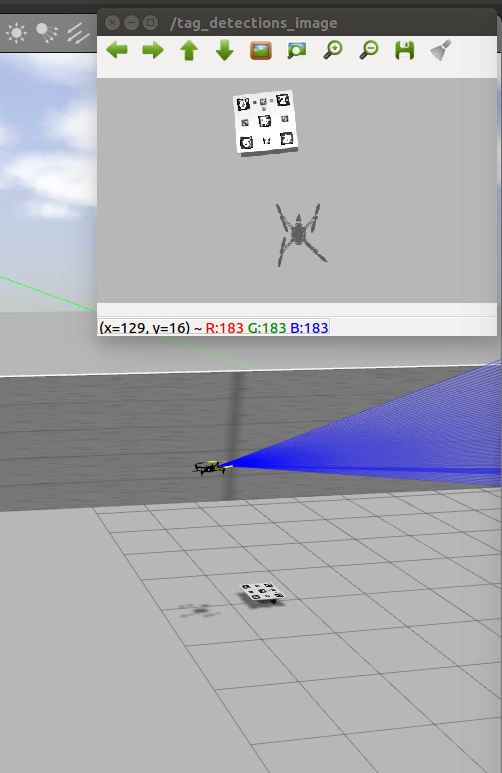}
  \caption{Simulation of drone tracking the moving platform. Inset is the simulated camera image as seen by the downward-facing camera on the drone.}
  \label{fig:simulation}
\end{center}
\end{figure}

All algorithms were tested first in simulation using Gazebo and Software in the Loop simulation. This realistically mimics the dynamics of the drone by using an accurate 3D model and the same firmware used onboard. Figure \ref{fig:simulation} shows a screenshot of the simulation as the algorithm was tested. The platform is being moved by a differential drive rover that carries the patterned board.

\begin{figure}[h]
  \includegraphics[width=\linewidth]{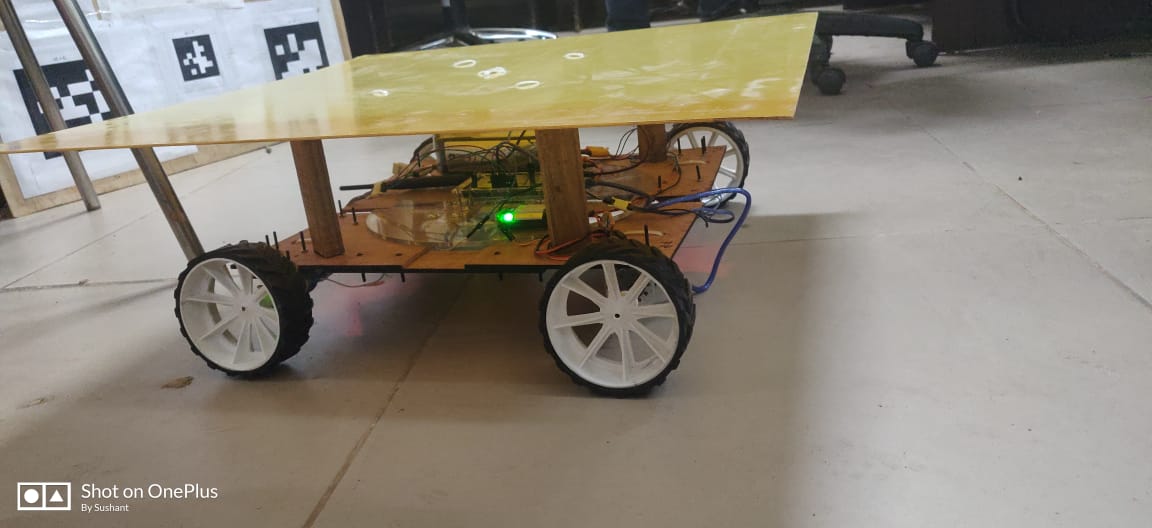}
  \caption{Four-wheel differential drive rover constructed to test the algorithm.}
  \label{fig:rover}
\end{figure}

\begin{figure}[h]
\begin{center}
  \includegraphics[width=0.5\linewidth]{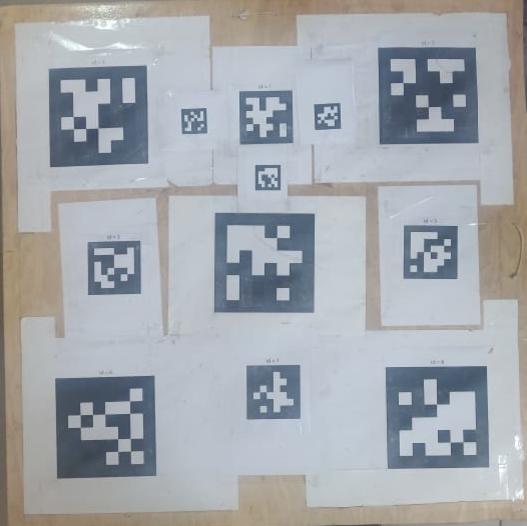}
  \caption{Marker board placed on the rover.}
  \label{fig:board}
\end{center}
\end{figure}

To test the system in actuality, a four-wheel differential drive rover, shown in Figure \ref{fig:rover}, capable of reaching speeds of up to 7kmph was constructed. The marker board, shown in Figure \ref{fig:board} was placed atop it, creating a 0.8m x 0.8m platform for the drone to land on.

The system was tested in challenging conditions with wind speeds varying between 15kmph to 30kmph and poor lighting due to constantly overcast weather. Both of these were factors that could potentially have affected our results, as vision-based navigation relies heavily on the clarity and contrast in the image, and precise landing requires stable conditions. However, as shown later, the algorithm performed well despite the conditions.

\section{Results}

In order to test the robustness and sensitivity of the algorithm, a simple test was created. The platform is placed behind the drone at the start and is brought into the frame of the camera after the drone has taken off. It moves at a constant speed along the global x-axis for a distance of 20m as the drone attempts to follow it at a height of 1m. The percentage of time in this period, from the first detection to the end of the experiment that the marker is detected is calculated, referred to here as tracking efficiency. This provides an accurate measure of the algorithm as continuous and highly efficient tracking is a necessity for precise landing. A higher tracking efficiency implies a more accurate landing. Table \ref{tb:pid_results} shows how tracking efficiency varies for different PID gains and platform speeds. These PID gains were set empirically by observing the response of the drone, although work is planned for a more systematic selection process. \emph{Note: Some values are left blank. These are gains for which the drone was unable to track the target.} 

\begin{table}[h]
\begin{center}
\label{tb:pid_results}
\caption{Tracking Efficiency at various PID gains for different platform speeds}
\begin{tabular}{c c c c c}
kp 	& ki & kd & Platform   & Tracking\\ 
	& 	 & 	  & Speed (m/s) & Efficiency \\\hline\\
0.50 & 0.0015 & 0.05 & 0.50 & 92.80\%\\
0.50 & 0.0015 & 0.05 & 0.75 & 51.04\%\\
0.50 & 0.0015 & 0.05 & 1.00 & 23.36\%\\\\

0.50 & 0.0015 & 0.07 & 0.50 & 89.19\%\\
0.50 & 0.0015 & 0.07 & 0.75 & 63.73\%\\
0.50 & 0.0015 & 0.07 & 1.00 & 34.03\%\\\\ 

0.50 & 0.0015 & 0.09 & 0.50 & 65.74\%\\ 
0.50 & 0.0015 & 0.09 & 0.75 & -\\
0.50 & 0.0015 & 0.09 & 1.00 & -\\\\ 

0.65 & 0.0015 & 0.05 & 0.50 & 74.48\%\\
0.65 & 0.0015 & 0.05 & 0.75 & 47.31\%\\
0.65 & 0.0015 & 0.05 & 1.00 & -\\\\

0.65 & 0.0015 & 0.07 & 0.50 & 87.45\%\\
0.65 & 0.0015 & 0.07 & 0.75 & 59.59\%\\
0.65 & 0.0015 & 0.07 & 1.00 & -\\\\ 

0.65 & 0.0015 & 0.09 & 0.50 & 93.35\%\\ 
0.65 & 0.0015 & 0.09 & 0.75 & 52.00\%\\
0.65 & 0.0015 & 0.09 & 1.00 & -\\\\  \hline
\end{tabular}
\end{center}
\end{table}

\begin{figure}[h]
  \includegraphics[width=\linewidth]{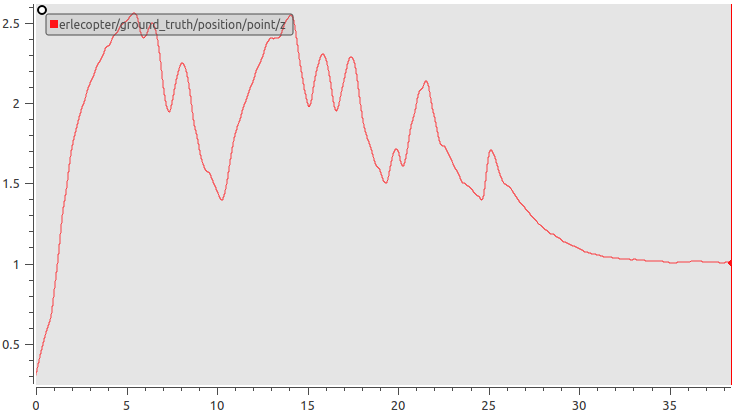}
  \caption{Plot of altitude of the drone vs time.}
  \label{fig:zvtime}
\end{figure}

This table shows some clear trends. For a given PID gain, as the speed of the target increases, the drone tracks it less efficiently. However, it is important to note that despite the fact that the tracking efficiency is low in some cases, the drone is still able to track the target. This is based on the assumption that all changes in speed and direction of the target are gradual, so the done continues to move in the same direction it was moving before it lost the detection, and it raises its altitude to increase the field of view, thus recapturing the target and resuming tracking. This motion is visible in Figure \ref{fig:zvtime} where it moves up, detects, and descends before finally settling into a steady state.

Another notable observation is that for higher proportional and derivative gains, although the tracking efficiency is similar to that of lower gains at small speeds, for higher speeds, it is unable to track the target. This is because of the inherent instability in the system. Although it initially tracks the platform, any slight non-linearities are magnified and the drone performs increasingly large oscillations above the target, eventually losing it. This tendency is much less at lower PID gains.

\begin{figure}[h]
  \includegraphics[width=\linewidth]{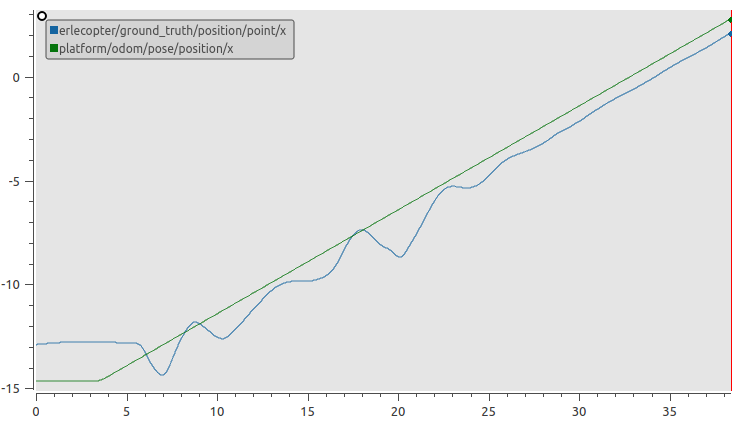}
  \caption{Plot of global x position of the drone and the platform vs time.}
  \label{fig:xvtime}
\end{figure}

\begin{figure}[h]
  \includegraphics[width=\linewidth]{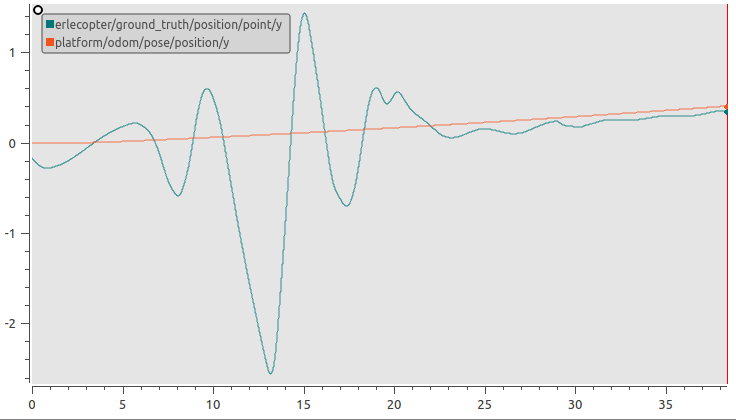}
  \caption{Plot of global y position of the drone and the platform vs time.}
  \label{fig:yvtime}
\end{figure}

Figure \ref{fig:xvtime} and \ref{fig:yvtime} also shows the variation along x and y of the drone with respect to time, with the straight line representing the position of the target. Tracking settles down to a steady state in about 25 sec.

\begin{figure}[h!]
  \includegraphics[width=\linewidth]{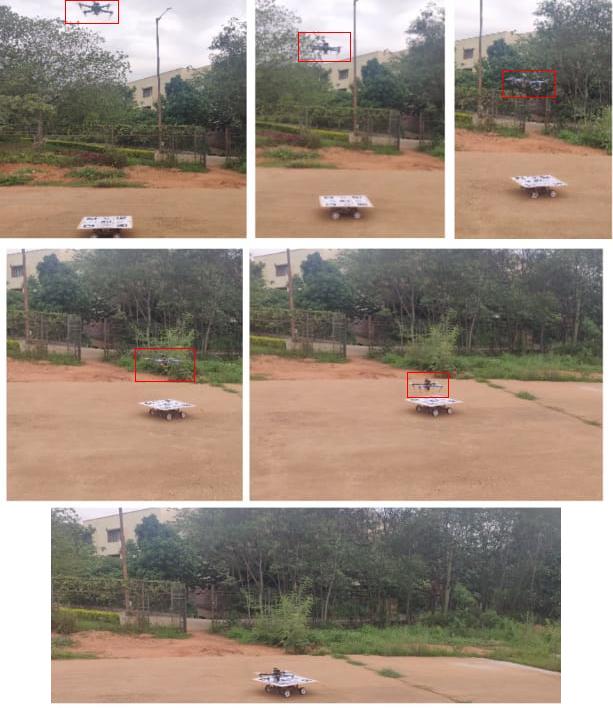}
  \caption{Frames showing the descent of the drone onto the target}
  \label{fig:landing}
\end{figure}

The system was also tested outdoors in actual flight and was able to land the drone on the rover while moving at a speed of 1.5m/s. The descent is shown in Figure \ref{fig:landing}.

\section{Future Work}
Although the algorithm works well once the PID gains are tuned, it fails when there are large changes in speed or direction. This can be solved by using a dynamic PID control, where the gains are set dynamically based on a velocity estimate of the target. This can be obtained using an Extended Kalman Filter (EKF) or Machine Learning. Both of these are areas we will be exploring in the future to improve the robustness of the system and increase the speed at which targets can be tracked and landed upon.

\begin{ack}
This work is possible due to the funding and support of the Wipro - IISc Research Innovation Network.
\end{ack}

\bibliography{ifacconf}{}
\nocite{*}
                                                   







\end{document}